\pdfoutput=1

\documentclass[11pt]{article}

\usepackage[preprint]{latex/acl}

\usepackage{times}
\usepackage{latexsym}
\usepackage{todonotes}

\usepackage[T1]{fontenc}

\usepackage[utf8]{inputenc}

\usepackage{microtype}

\usepackage{inconsolata}

\usepackage{graphicx}

\newcommand{\mk}[1]{}
\newcommand{\mc}[1]{}

\title{An Interdisciplinary Approach to Human-Centered Machine Translation}

\author{
 \textbf{Marine Carpuat\textsuperscript{1}},
 \textbf{Omri Asscher\textsuperscript{2}},
 \textbf{Kalika Bali\textsuperscript{3}},
 \textbf{Luisa Bentivogli\textsuperscript{4}},
  \textbf{Fr\'ed\'eric Blain\textsuperscript{5}},  
 \textbf{Lynne Bowker\textsuperscript{6}},\\
 \textbf{ Monojit Choudhury\textsuperscript{7}},
 \textbf{Hal Daum\'e III\textsuperscript{1}},
 \textbf{Kevin Duh\textsuperscript{8}}, 
 \textbf{Ge Gao\textsuperscript{1}}, 
  \textbf{Alvin Grissom II\textsuperscript{9}}, \\ 
 \textbf{Marzena Karpinska\textsuperscript{3}},
  \textbf{Elaine C. Khoong\textsuperscript{10}},
 \textbf{William D. Lewis}\textsuperscript{11},
\textbf{André F. T. Martins\textsuperscript{12}}, \\
 \textbf{Mary Nurminen\textsuperscript{13}},  \
 \textbf{Douglas W. Oard\textsuperscript{1}},  
 \textbf{Maja Popovic\textsuperscript{14}}, 
 \textbf{Michel Simard\textsuperscript{15}},
 \textbf{François Yvon\textsuperscript{16}}
\\
\small{
 \textsuperscript{1}University of Maryland,
 \textsuperscript{2}Bar-Ilan University,
 \textsuperscript{3}Microsoft, 
 \textsuperscript{4}Fondazione Bruno Kessler, 
 \textsuperscript{5}Tilburg University,  
\textsuperscript{6}Université Laval, } \\
\small{
 \textsuperscript{7}Mohamed bin Zayed University of Artificial Intelligence,
 \textsuperscript{8}Johns Hopkins University,
 \textsuperscript{9}Haverford College, }\\
 \small{
 \textsuperscript{10}University of California, San Francisco, 
\textsuperscript{11}University of Washington, 
 \textsuperscript{12}Instituto Superior Técnico, Universidade of Lisboa } \\
\small{
\textsuperscript{13}Tampere University, 
 \textsuperscript{14}Dublin City University \& IU University, 
 \textsuperscript{15}National Research Council Canada 
 \textsuperscript{16}Sorbonne-Université \& CNRS }\\
 \small{
   \textbf{Correspondence:} \href{mailto:marine@umd.edu}{marine@umd.edu}
   }
}

\begin{document}
\maketitle
\begin{abstract}
Machine Translation (MT) tools are widely used today, often in contexts where professional translators are not present. Despite progress in MT technology, a gap persists between system development and real-world usage, particularly for non-expert users who may struggle to assess translation reliability.
This paper advocates for a human-centered approach to MT, emphasizing the alignment of system design with diverse communicative goals and contexts of use. We survey the literature in Translation Studies and Human-Computer Interaction to recontextualize MT evaluation and design to address the diverse real-world scenarios in which MT is used today.

\end{abstract}

\mc{Other potentially relevant papers: \citep{riley-etal-2023-frmt,wang-etal-2023-controlling}}

\section{Introduction}

Machine Translation (MT) is one of the few NLP technologies that has been widely available online for decades. As both translation quality and internet access have improved \citep{gaspari-hutchins-2007-online}, MT has gained a large and diverse user base.
Millions of people use it to communicate across languages, including in settings where professional translators or interpreters are not realistically available
 \citep{nurminen-papula-2018-gist,kaspere-etal-2021-sustainable,vieira-etal-2022-machine,kenny-etal-2022-machine}.

As MT becomes increasingly embedded in everyday tools and tasks, the socio-technical gap between how the technology is developed and how it is used in real-world contexts is widening \citep{ackerman-2000-intellectual}. Whereas initial MT systems were primarily used to support professional translators or narrow domains \citep{hutchins-2001-machine}, today MT can be used by anyone with internet access in their daily life \citep{yvon-2019-two,kenny-etal-2022-machine}. However, MT does not yet fulfill its promise to enable communication across languages, particularly for users who may lack the language or domain expertise needed to make informed use of the translations \citep{liebling-etal-2020-unmeta,santy-etal-2021-language,valdez-etal-2023-migranta}. This gap is further amplified by the rise of translation with general-purpose large language models (LLMs) \citep{vilar-etal-2023-prompting,alves-etal-2024-tower,kocmi-etal-2024-findings,hendy2023goodgptmodelsmachine}. With such tools, translation can be integrated into broader workflows, where translation might be covert, making it even harder for users to assess its reliability. This can result in over-trust in MT \citep{martindale-carpuat-2018-fluency}, which is particularly problematic in high-stakes scenarios where it can cause harm \citep{VieiraOHaganOSullivan2021}, but also in under-use of MT tools in cases where they could be beneficial \citep{obrien-federici-2019-crisis}. 

We argue that a human-centered approach to MT is needed: one that broadens what MT systems do to help users weigh risks and benefits and align system design with communicative goals. This approach echoes calls for human-centered AI \citep{capel-brereton-2023-what}, which includes recognizing that people are at the heart of the development of any AI system \citep{vaughan-wallach-2021-humancentered},  emphasizing designing AI systems that augment rather than replace human capabilities, prioritizing human agency and system accountability \citep{shneiderman-2022-humancentered}, %
 and using human-centered design methods for AI systems \citep{chancellor-2023-practices}.

To provide a foundation for human-centered MT, we argue that it is important to adopt an interdisciplinary approach that includes Translation Studies and Human-Computer Interaction (HCI). In this paper, we recontextualize  MT research by surveying relevant literature in these fields.  
As \citet{GreenHeerManning2015} point out, the question of how to design effective human–MT interaction has been considered long before HCI, NLP, or AI were formalized disciplines. For example, %
\citet{kay-1980-proper} introduced a cooperative interactive system as an alternative to fully automated translation to replace professional translators. As MT improved, these questions were revisited to design mixed-initiative post-editing interfaces \citep{GreenHeerManning2013,koehn-etal-2014-casmacat,briva-iglesias-etal-2023-impact}, highlighting the benefits of designing MT systems to augment, rather than replace, professional translators’ abilities \citep{obrien-2024-humancentered}.
As the MT user base has expanded from professional translators to professionals in other disciplines, as well as the general public \citep{savoldi-etal-2025-translation}, many relevant lessons can be drawn from theoretical and empirical work in Translation Studies and HCI. 
Accordingly, this survey results from discussions between co-authors across these fields. Translation studies and HCI experts identified key insights they wished to share with the MT researchers. These insights served as points of connection with the MT literature.

Considering MT's diverse uses (Section~\ref{sec:uses}), we synthesize cross-disciplinary insights spanning MT literacy (Section~\ref{sec:mtliteracy}), human-MT interaction (Section~\ref{sec:mtoutside}), and translation ethics (Section~\ref{sec:ethics}). We then outline research directions for human-centered MT evaluation (Section~\ref{sec:hceval}) and design (Section~\ref{sec:hcmt}), illustrating interdisciplinary human-centered MT research with a healthcare case study (Section~\ref{sec:casestudy}).

\section{Understanding Contexts of Use}
\label{sec:uses}

To develop human-centered MT, we must first understand how MT is used in the real world. While the body of research on users, contexts, and purposes has shown increased growth recently, the considerable size of the user population, estimated in 2021 at more than one billion \citep[p. 23]{nurminen-2021-investigatinga}, and growing variety of use contexts present a challenge for synthesizing that research into knowledge that can be used for designing systems that more directly serve user needs.

A classical framework distinguishes three use types \citep{HovyKingPopescu-Belis2002a}: assimilation, in which MT helps users get the gist of content in a foreign language (e.g., browsing news, triage) without requiring perfect quality; dissemination, in which MT content is shared with others, demanding higher quality (e.g., public announcements); and communication, in which MT supports live or interactive multilingual exchanges (e.g., chat, classrooms).

A wealth of MT research projects have considered different use cases over the years, but without much information sharing across settings: classroom speech translation \citep{lewis-niehues-2023-automatic}, healthcare \citep{khoong-etal-2019-assessing,valdez-guerberof-arenas-2025-google}, crisis response \citep{lewis-etal-2011-crisisa,escartin-moniz-2019-ethical}, international patent processes \citep{nurminen-2020-raw}, migration scenarios \citep{vollmer-2020-digital,vieira-2024-machine,pieta-valdez-2024-migration}, research and academic writing  \citep{bowker-ciro-2019-machine,ehrensberger-dow-andlehr-2023-new,bawden-etal-2024-translate}, customer support \citep{goncalves-etal-2022-agent}, literary MT \citep{karpinska-iyyer-2023-largea, zhang-etal-2025-how}, and CAT/localization \citep{koehn-etal-2014-casmacat, lin-etal-2010-composing}, and intercultural collaboration platforms \citep{ishida-2016-intercultural}. The examination of these contexts of use alone suggests some considerations that should impact MT design, beyond the general purpose of translation: risk management (error tolerance varies by domain), synchrony (real-time vs. delayed), urgency, shelf life, audience, interaction dynamics, modality/accessibility, and overtness of MT use (e.g., covert use of MT on a multilingual website or embedded in another application). 

We also lack a deeper understanding of who uses online MT tools and how. \citet{nurminen-2021-investigatinga} estimates that 99.97\% of MT users are not professional translators. “Machine Translation Stories” illustrate diverse uses by individuals from all walks of life, from music students translating old Italian arias to people using MT in their professional life \citep{nurminen-2021-machine}. A survey of 1,200 UK residents shows high satisfaction with MT for low-stakes uses but highlights a demand for better quality \citep{vieira-etal-2022-machine}. 
Another survey of 2,520 UK public service professionals reveals that 33\% had used MT in their work, predominantly within health and social care sectors, but also across legal, emergency, and police services \citep{nunesvieira-2024-uses}. Formal training was uncommon, leading many professionals to rely on personal devices and publicly available tools like Google Translate and ChatGPT.
But user needs are not met equally across socioeconomic and geographic contexts. For instance, interview studies showed that MT applications do not support effective cross-lingual communication for migrant workers in India and immigrant populations in the U.S., resulting in significant negative impacts on their daily lives \citep{liebling-etal-2020-unmeta}.

Human-centered MT should not just respond to user needs \citep{gasson-2003-humancentered}, but consider more broadly how people are affected by MT, including the languages and perspectives of marginalized populations \citep{bender2024power},
and considering both direct and indirect stakeholders \citep[p. 39]{friedman-hendry-2019-value}. These include recipients of translated content, institutions using MT at scale \citep{koponen-nurminen-2024-risk}, writers of source texts \citep{taivalkoski-shilov-2019-ethical,lacruzmantecon-2023-authorship}, MT practitioners \citep{robertson-etal-2023-angler}, language learners, and broader language communities given evidence that language evolves through automation \citep{guo-etal-2024-curiousa}.

This complexity calls for further investigation of MT in context and for organizing use cases into a taxonomy that balances general-purpose development with contextual needs. %

\section{Machine Translation Literacy}
\label{sec:mtliteracy}

Translation Studies research highlights a need for promoting machine translation literacy \citep{bowker-ciro-2019-machine} given the wide gap between how translation is approached by people within versus beyond the language professions.
Professional translators have been trained in translation, which usually also involves acquiring a domain specialization \citep{scarpa-2020-research}, such as legal, medical or technical translation. As people, professional translators also have deep knowledge of the language pair in question, and the type of real-world knowledge and cultural knowledge that is necessary when translating between languages and cultures. %
Translators can bring all this information to bear on their understanding of the source text. They compensate for shortcomings in the source text (e.g., they can clarify the intended meaning of a sentence with poor punctuation or where a homophone has erroneously been used). Professional translators also operate within a sort of decision-making framework because they request (or even require) a translation brief from their client or employer \citep{munday-etal-2022-introducing}. The translation brief is essentially a set of instructions and information that helps the translator to make sensible choices. For instance, the brief contains information about the intended purpose of the translation, where it will be published, who will read or use it, what the target reader’s background (language variety, culture, education level) is. All of this information allows the translator to make informed decisions. 

In contrast many MT users have no background in translation. They may not have the necessary linguistic knowledge, domain or cultural knowledge required to evaluate the adequacy of the translated text. They may have misconceptions about translation \citep{bowker-2023-demystifying}, e.g., seeing it as an exact science or a task that can be done by any bilingual. They might not realize the importance of the translation brief. In short, they lack MT literacy, which has been defined as ``knowing how MT works, how the technology can be useful in a particular context, and what the implications are of using it for various purposes'' \citep{obrien-ehrensberger-dow-2020-mta}. 

This highlights the necessity of MT literacy and motivates a key direction in Human-Centered MT: designing tools that promote informed and responsible use, especially by lay users. Current tools lack this, but we will see that the existing literature provides a starting point.

\section{Empirical Studies of MT Outside Professional Translation} 
\label{sec:mtoutside}

Translation Studies and HCI offer extensive empirical research on human-MT interaction within various contexts, beyond professional translation. It reveals existing user strategies for using potentially imperfect MT, interventions that have already shown promise, and open research directions.

\paragraph{Post-editing} The most studied human-MT interaction setting is probably post-editing, where people edit raw MT to improve it. It has received significant attention in the context of professional translation \citep[among others]{cadwell-etal-2016-human,briva-iglesias-etal-2023-impact}, but it is also performed by other users,  for instance when they translate their own source text as a writing aid in academic settings \citep{bowker-2020-chinese,xu-etal-2024-surveya,obrien-etal-2018-machine} or for scientific dissemination \citep{bawden-etal-2024-translate}. 
There is evidence that even monolingual users can interpret and revise MT output when provided with background knowledge or translation options \citep{hu-etal-2010-translation,koehn-2010-enabling}.%

When users do not understand the target language, post-editing is not an option, but they still face a decision about whether to publish or share the raw MT outputs. \citet{zouhar-etal-2021-backtranslation} studies the impact of augmenting raw MT with backtranslation, source paraphrasings and quality estimation feedback in such ``outbound translation'' settings, and show that backtranslation feedback increases user confidence in the produced translation, but not the actual quality of the text produced.

\paragraph{Augmented Outputs for Gisting} Several studies show that augmenting MT outputs can improve comprehension and engagement, particularly when MT is used for understanding the gist of a text.
Highlighting key words in source and target texts can improve people's ability to understand  difficult translations \citep{pan-wang-2014-enhancing,grissom-ii-etal-2024-rapidly}, and adding emotional and contextual cues promotes engagement with social media posts in a foreign language \citep{lim-etal-2018-translation}.  

Research has also shown that users sometimes access outputs from multiple MT tools to better understand the errors associated with each individual output and, in doing so, enhance overall comprehension \citep{anazawa-etal-2013-use,nurminen-2019-decisionmaking,robertson-etal-2021-three}. Other research has also indicated positive effects from exposure to outputs from multiple MT tools \citep{xu-etal-2014-improving,gao-etal-2015-twoa}. Human-centered tools for MT gisting might therefore involve MT tools that embed a second MT tool directly into their user interface \citep{nurminen-2020-raw}, or perhaps automatically show two outputs as a low-cost means of enhancing users' perceived transparency. 
\mc{add Marianna Martindale's MT Summit 2025 paper once published}

\paragraph{Source Understanding}
People use MT not only to gain access to texts across language boundaries, but also to augment and ensure their understanding of texts that are in languages they have limited competence in \citep{nurminen-2021-investigatinga}. They might position a source text and its translation side-by-side and refer to both while reading, or they may look at both original and translated messages in an MT-mediated conversation \citep{nurminen-2016-machine}. Recognizing this tendency, human-centered MT tools could make it easy to access original texts alongside their machine-translated versions, and provide affordances to compare them easily.

\paragraph{MT-mediated Communication} HCI research has studied MT-mediated communication, and how the use of MT affects not only performance, but also interpersonal dynamics. Empirical evidence shows that people develop their own strategies to compensate for imperfect MT, such as adapting what they say (e.g., by employing redundant expressions and suppressing lexical variation in language use) \citep{yamashita-ishida-2006-effectsa,hara-iqbal-2015-effecta}, using back-translation to assess outputs in a language they do not understand \citep{ito-etal-2023-use}, or simply relying on their holistic understanding of the conversation to fill in gaps where the MT output does not make sense \citep{robertson-diaz-2022-understanding}.
Even when effective, these strategies come at a cost to communication: people communicate less naturally and authentically \citep{yamashita-ishida-2011-conversational} and might get misleading signals on translation quality \citep{tsai-wang-2015-evaluating}. Imperfect translations also affect interpersonal dynamics between interlocutors, increasing the risk of participants misinterpreting their task partner's intent \citep{lim-etal-2022-understandingb}, misattributing communication breakdowns to human vs. MT-generated errors \citep{10.1145/2531602.2531702, robertson-diaz-2022-understanding}, and misassessing one another's contribution to the collaborative task \citep{10.1145/3686942}.%

\paragraph{Trust}
Lay users' trust in MT is largely shaped by their perception of how MT-as-a-black-box functions, not just its intrinsic quality. Identical translations can be perceived differently when labeled machine vs. human-generated \citep{AsscherGlikson2021}, and
people might assign inconsistent ratings to the MT outputs before vs. after the label is disclosed \citep{bowker-2009-can}. 
Not all MT errors impact user trust equally: fluency or readability errors tend to lower trust more than adequacy errors, even though the latter can be more misleading when users rely on MT-generated meanings to inform their actions \citep{martindale-carpuat-2018-fluency, popovic-2020-relations}. Factors like language proficiency, subject knowledge, and MT literacy influence how users perceive MT quality in gisting contexts \citep{nurminen-2021-investigatinga}.
MT literacy 
has also been shown to play a significant role in shaping translators’ trust in MT \citep{scansani-etal-2019-translator}.

Taken together, this body of work highlights that building truly human-centered MT systems demands much more than generating fluent and adequate translations. It requires aligning system design with real-world communication practices, developing interaction strategies that empower users, and supporting their ability to assess risks in ideally independent and time-sensitive ways. Crucially, it also means empirically studying how these systems affect stakeholders, not just in terms of task performance, but also in how they shape interpersonal dynamics and shared understanding.%

\section{Ethicality of MT}
\label{sec:ethics}

The social implications of MT use extend beyond its immediate usefulness, bringing us into the realm of ethics.
What does it mean for MT to be ethical?
Surveying major frameworks of translation ethics  \citep{koskinen-pokorn-2020-routledge} provides a foundation for addressing this question, highlighting the inherent multiplicity and conflicting perspectives in determining what is right or wrong in practice \citep{chesterman-2001-proposal,lambert-2023-translation}.

For example, some approaches base translation ethics on strictly representing the original text's ``precise'' meaning and form, at all costs and under all circumstances \citep{newmark-1988-textbook}. Other approaches emphasize a functional ethics of service,  where ethical translation is defined by the translator's adherence to the requester's instructions, even if this means changing the source text or using it as mere inspiration \citep{holz-manttari-1984-translatorisches}. Others prioritize alterity and social justice, viewing translation as a tool to challenge social and political inequalities by reframing communities' identity and values; in this case, ethical action might even involve refusing to translate the source text \citep{robinson-2014-translation}. 
Several other translation ethics frameworks exist, each revolving around different priorities and values \citep{koskinen-pokorn-2020-routledge}.

Today's influential ethical frameworks also imply that the translator’s ethical response is necessarily situation- and text-dependent \citep{pym-2012-translator}. By this we mean that for different texts, and in different situations, the ethical decision – whichever ethical framework one follows – may take different shapes. MT ethics, then, are no less situation-contingent than issues of MT usability or effectiveness.

Finally, a typology of the main approaches to translation ethics also reveals how some ethics are largely regional, or field-specific, inasmuch as they stem from the particular features of translation as a medium for intercultural communication \citep[p. 57]{pym-2012-translator}. In contrast, other approaches are more general in their concerns and values, and not intrinsic to the field of translation as such. %
Along these lines, it could be argued that a useful implementation of human-centered ethical evaluation in the case of MT should involve the compartmentalization of MT ethics from general AI ethics, and the preference for regional frameworks of ethics for MT \citep[p. 102--109]{asscher-2025-machine}. This %
implies a give-and-take %
between MT ethics oriented to the specificities of translation, on the one hand, and universal ethics, reminiscent of the general protocols of AI ethics proposed so abundantly in recent years, on the other hand \citep{floridi-etal-2018-ai4peoplean}.

Relating these ethical insights to MT can apply to both the increasingly autonomous decision-making of the tool itself, and the social conditions that underpin its development and maintenance  \citep[p. 98--101]{asscher-2025-machine}. %
The development and use of MT has already had vast consequences for many stakeholders. %
The ownership and distribution of anonymized translation data needed for the development of MT systems, and the re-use of this data to fine-tune MT, are some of the issues at stake, as there is currently no compensation for the original human translators who created the data, and MT systems serve causes that are opaque to these translators and might contradict their values \citep[p. 123--126]{moorkens-2022-ethics}. Issues of confidentiality and privacy are also pertinent, as personal translation data is utilized to train MT systems without regulation, rendering this data potentially identifiable \citep{nunesvieira-etal-2022-privacy}. The risks involved in high-stakes use of MT may strain the question of the moral and legal responsibility even further, for example in medical and legal situations, where translation errors may be particularly consequential \citep{VieiraOHaganOSullivan2021}. Then, there are the sometimes problematic uses of MT in the professional translation workflow, and the broader issues of sustainability of the translation industry and environmental concerns %
\citep{bowker-2020-translation,skadina-etal-2023-deep,shterionov-vanmassenhove-2023-ecological}. MT ethics also apply to the cultural and gender bias of contemporary LLMs \citep{gallegos-etal-2024-bias}, which may be manifested in translation, or the censorship recently enacted in some generative AI tools concerning certain charged historical occurrences, reinforcing unequal power relations across cultures \citep{wang-etal-2025-how,bianchi-etal-2023-gender}.

Considering these points, human-centered MT research must pursue richer assessments of the moral consequences of its use in society. Studies of MT ethicality are valuable regardless of immediate implementability and can inform business and scientific leadership in governing the field and shaping MT agency and social implications.

\section{Human-Centered MT Evaluation}
\label{sec:hceval}

MT evaluation has focused on benchmarking systems, or rating  individual outputs, using automatic or human ratings of translation quality as ground truth \citep{white-oconnell-1993-evaluation,koehn-monz-2006-manual,graham-etal-2013-continuous,LaubliCastilhoNeubigSennrichShenToral2020,freitag-etal-2021-expertsa}. Some recent proposals call for broadening its scope to measure social and environmental impact in addition to performance \citep{moorkens-etal-2024-proposal,santy-etal-2021-language}. A human-centered approach can draw from conceptualizations of the translation process and product quality from Translation Studies \citep{liu-etal-2024-evaluation}, and HCI methodology for evaluating systems in their socio-technical context \citep{liebling-etal-2022-opportunities}.

\paragraph{From Generic to Situated MT}
A key shift is from generic, context-independent evaluation toward situated assessments of fitness-for-purpose and stakeholder impact. %
Holistic quality scores \citep{graham-etal-2013-continuous} are already complemented by fine-grained annotations such as MQM \citep{lommel-etal-2014-multidimensional}. 
In contrast, Translation Studies work emphasizes evaluating translations based on their suitability for their intended purpose rather than adhering to a one-size-fits-all notion of quality \citep{bowker-2009-can,chesterman-wagner-2014-can,colina-2008-translationa}. The impact of MT errors thus needs to be assessed in context~\citep{agrawal2024assessing}, as general benchmarks may obscure rare but extreme errors~\citep{shi-etal-2022-rare}. Expert knowledge might be required, for instance to determine whether an adequacy error poses a clinical risk \citep{khoong-etal-2019-assessing}, or to assess social harms such as gender bias \citep{savoldi-etal-2021-gender, savoldi-etal-2024-what}, name mistranslation \citep{sandoval-etal-2023-rose}, and lack of cultural awareness \citep{yao-etal-2024-benchmarkinga}. Providing an ``evaluation brief'' \citep{liu-etal-2024-evaluation} can describe the circumstances surrounding the translation creation, who it is for, and how it is intended to be used. Evaluation through question answering is another way to assess if translations preserve important information \citep{ki2025askqe,fernandes2025llms}.

\paragraph{From Annotation to Human Studies} Human studies that incorporate MT within the relevant end-user task can help us assess the impact of MT more comprehensively.
Such tasks might align closely with the production and understanding of translations, such as post-editing MT \citep{castilho-obrien-2016-evaluating,castilho2017acceptability,bawden-etal-2024-translate,savoldi-etal-2024-what}, reading comprehension \citep{JonesGibsonShenGranoienHerzogReynoldsWeinstein2005,scarton-specia-2016-reading}, gisting \citep{nurminen-2021-investigatinga} or triage tasks \citep{martindale-carpuat-2022-proposeda}.
MT might be a tool in support of another task, such as collaborative information exchange in teams \citep{yamashita-ishida-2006-effectsa}, social media consumption \citep{lim-etal-2018-translation}, hiring and personnel decision making \citep{zhang-etal-2022-facilitatingb} or housing information seeking \citep{xiao-etal-2025-sustaining}, and everyday conversations \citep{robertson-diaz-2022-understanding}. As~\citet{santy-etal-2021-language} show, in such real-world cases, machine-aided translation systems can bring significant value to end-users. Nevertheless, this value is often contextualized within trade-offs among time, performance, and computational cost, especially given the limited technical accessibility and important occurrence of low-resource language settings.

\paragraph{From Static Benchmarks to Iterative Design} Evaluations with human users do not only occur at the end of a project; rather, they drive the iterative refinement cycle of the entire human-centered design process. This process typically begins with needs-finding studies to identify the social problem that technical solutions aim to resolve \citep{gao-fussell-2017-kaleidoscope, gao-etal-2022-taking, 10.1145/3686942}. It is often followed by co-design activities, where existing tools are used as technology probes to elicit inputs from targeted user groups on MT design. Subsequent phrases include usability testing or clinical trials after each round of system development to determine the degree of success \citep{khoong-rodriguez-2022-research}. 
There exist a wealth of frameworks to guide this process, including Human-centered design, Participatory Design, and Value Sensitive Design \citep{friedman-hendry-2019-value}, all of which foreground the values of direct and indirect stakeholders. MT evaluation can also draw from frameworks for trustworthy AI, particularly methods for studying mental models \citep{bansal-etal-2019-accuracy}, trust calibration \citep{vereschak-etal-2021-how}, and how a human-AI work system performs \citep{hoffman-etal-2023-measures}. These efforts aim to ensure that MT systems can account for the complex dynamics between system outputs, user interpretations, and downstream consequences, thereby requiring interdisciplinary collaborations and tailored study designs.

\section{Human-Centered MT Design}
\label{sec:hcmt}

This section outlines emerging techniques that can reframe MT as a contextual, potentially interactive process responsive to users' needs, moving beyond traditional sequence transduction. It provides a richer toolbox to support MT literacy (Section~\ref{sec:mtliteracy}) and build on past empirical studies of human-MT interaction  (Section~\ref{sec:mtoutside}).

\paragraph{Richer Inputs, Many Outputs}
Human-Centered MT must adapt outputs to the audience and context. Research has already explored controlling formality \citep{sennrich-etal-2016-controllinga,rippeth-etal-2022-controlling}, style \citep{niu-etal-2017-study,agarwal-etal-2023-findings}, complexity \citep{agrawal-carpuat-2019-controlling,oshika-etal-2024-simplifying}, and personalization \citep{MirkinMeunier2015,rabinovich-etal-2016-personalized}. Adaptation may also require explaining content \citep{srikanth-li-2021-elaborative,han-etal-2023-bridging,saha-etal-2025-reading}, or warning about cultural misunderstandings \citep{pituxcoosuvarn-etal-2020-effect,yao-etal-2024-benchmarkinga}. However, it is still unclear how users and other stakeholders can guide these systems in proactive and ecologically valid ways.

More contextual inputs are needed, similar to translator briefs \citep{castilho-knowles-2024-survey}. MT work has considered incorporating domain knowledge \citep{ClarkLavieDyer2012,chu-wang-2018-survey}, style labels \citep{sennrich-etal-2016-controllinga,niu-etal-2017-study}, example translations \citep{xu-etal-2023-integrating,agrawal-etal-2023-incontext,bouthors-etal-2024-retrieving}, and terminology \citep{alam-etal-2021-findings,michon-etal-2020-integrating}. Some also address long-form \citep{karpinska-iyyer-2023-largea,peng-etal-2024-propos} and conversational translation \citep{BawdenBilinskiLavergneRosset2021,pombal-etal-2024-contextaware}. However, these efforts usually consider one dimension of context at a time; we still need more holistic approaches that take a broad view of context  \citep{castilho-knowles-2024-survey} and incorporate knowledge and feedback needed for culturally appropriate outputs \citep{tenzer-etal-2024-ai,saha-etal-2025-reading}.

\paragraph{An Iterative Translation Process}
LLMs enable multi-stage translation workflows, including pre-editing, evaluation, and post-editing \citep{briakou-etal-2024-translating,alves-etal-2024-tower}. Pre-editing involves rewriting source texts to improve MT output \citep{bowker-ciro-2019-expanding,stajner-popovic-2019-automated,ki-carpuat-2025-automatic}, while post-editing—either human or automatic—is  studied widely \citep{lin-etal-2022-automatic,vidal-etal-2022-automatic,ki-carpuat-2024-guidinga}. Yet, most work remains system-centric. Interactive approaches designed for professional translators \citep{GreenHeerManning2013,briva-iglesias-etal-2023-impact} suggest benefits from involving lay users with diverse goals and levels of proficiency.

\paragraph{Scale \& Context} How can we specialize models for specific contexts while reaping the benefits of scale \citep{team-etal-2022-no,johnson-etal-2017-google`s,vilar-etal-2023-prompting,kocmi-etal-2024-findings}? Work in this direction could build on efforts to structure resources for horizontal (across languages) and vertical (across domains) generalization \citep{ishida-2006-language,rehm-2023-european}, and techniques to support task \citep{ye-etal-2022-eliciting,alves-etal-2024-tower}, language \citep{blevins-etal-2024-breaking}, and domain and terminology \citep{segonne-etal-2024-jargon} specialization in LLMs. 

\paragraph{Decentering MT}
Centering people means recognizing that MT is often just one part of a broader workflow, where the MT output is not the end product.
MT today often participate in content co-production with humans, rather than only for source-to-target conversion. This can be done via synchronized bilingual writing \citep{crego-etal-2023-bisynca, 10.1145/3686942} or using translation as an aid for scientific writing \citep{o2018machine,steigerwald-etal-2022-overcominga,ito-etal-2023-use}. In those settings, even when translating an abstract, the translation might be more of an adaptation than a literal translation \citep{bawden-etal-2024-translate}.
Translation can be implicit or partial, when supporting simultaneous interpreters~\citep{grissom-ii-etal-2024-rapidly}, enabling natural translanguaging practices of bilinguals \citep{10.1145/3706598.3714050}, or searching for texts written in a foreign language given a native language query \citep{galuscakova-etal-2022-crosslanguage,nair-etal-2022-transfer}. In those settings, human-MT interface design is critical for lay users to remain aware of features of the targeted content and to develop strategies for navigating it \citep{petrelli-etal-2006-which}.
The need for intelligent interface design is particularly pronounced in LLM-powered multilingual communication and user interactions with conversational agents, where models must interpret and generate content for fluid language use while adapting to user goals, styles, and cultural norms. 
To support this, a prompt engineering playground with customized MT and user interfaces may enhance the accessibility of LLMs for a broader population \citep{mondshine-etal-2025-english}.

\paragraph{Risk Management}
Reliable MT should help users weigh the benefits of MT against the risks it may pose. Quality estimation techniques designed for explainability have provided a good foundation toward this goal \citep{fomicheva-etal-2021-eval4nlp,guerreiro-etal-2023-xcomet,briakou-etal-2023-explaining, specia-etal-2018-quality}. That said, growing evidence from user studies shows that more work is needed to identify and assess risks~\citep{koponen-nurminen-2024-risk}, generate actionable feedback in user-specified contexts \citep{zouhar-etal-2021-backtranslation,mehandru-etal-2023-physiciana}, determine when and how to disclose the use of MT \citep{simard-2024-position, 10.1145/3686942},  provide useful descriptions of model properties \citep{mitchell-etal-2019-model}, promote MT literacy among lay users \citep{bowker-ciro-2019-expanding}, and support the development of accurate user mental models \citep{bansal-etal-2019-accuracy}. Frameworks from human-centered explainable AI, such as seamful design \citep{ehsan-etal-2022-seamful}, can help pinpoint gaps between system affordances and the needs of human stakeholders, fostering better alignment.

In sum, while existing work offers a rich toolbox for human-centered MT, more research is needed on designing interactions that preserve user agency and support effective, trustworthy use. This includes new interfaces that balance simplicity and flexibility, and foundational work on training models for controllability and context-awareness.

\section{Case Study: Toward Reliable Translation for Clinical Care}
\label{sec:casestudy}

Research on MT for clinical settings illustrates how human studies can drive the cycle of human-centered MT (Section~\ref{sec:hceval}) by understanding specific contexts of use (Section~\ref{sec:uses}) to guide interface and model design decisions (Sections~\ref{sec:mtoutside},\ref{sec:hcmt}).

\paragraph{Understanding Needs} Language barriers are a major source of healthcare disparities \citep{cano-ibanez-etal-2021-physicianpatient}, yet access to professional interpreters remains limited \citep{flores-2005-impact,ortega-etal-2023-language}. MT can potentially support clinical care, but reliability is a critical concern: MT errors can cause serious harm in, for example, discharge instructions from emergency departments \citep{khoong-etal-2019-assessing,taira-etal-2021-pragmatic}, pediatric care \citep{brewster-etal-2024-performance} or urology \citep{rao-etal-2024-comparative}, with disparate impact across languages. 
Yet, MT frequently mediates interactions between healthcare providers and patients in practice \citep{genovese-etal-2024-artificial}. While dedicated MT tools have been developed for clinical settings \citep{starlander-etal-2005-practicing,bouillon-etal-2005-generic}, generic apps such as Google Translate are still most commonly used \citep{nunesvieira-2024-uses}. In face of challenges such as time constraints, cultural barriers, and medical literacy gaps, clinicians develop their own workarounds when using MT, such as back-translation or relying on non-verbal cues to assess understanding \citep{mehandru-etal-2022-reliablea}.

\paragraph{Research Directions} Generic MT tools thus often fall short in clinical care, and needs-findings studies motivate research into integrating pre-translated medical phrases, multimodal communication support, and interactive tools to assess mutual understanding.
A human study evaluated feedback mechanisms to assist physicians in assessing the reliability of MT outputs in clinical settings, finding that quality estimation tools generally improve physicians' reliance on MT but fail to detect the most clinically severe errors \citep{mehandru-etal-2023-physiciana}. %
Complementary efforts focus on developing custom MT approaches that prioritize reliability and verifiability, by using vetted canonical phrases to scaffold the translation \citep{bouillon-etal-2017-babeldr} or guide users in crafting better MT inputs \citep{robertson-2023-designing}. While these works focus on text-based MT, many healthcare use cases also warrant consideration of interaction using speech \citep{spechbach-etal-2019-speechenabled}, sign language \citep{esselink-etal-2024-exploring} and pictographs \citep{gerlach-etal-2024-concept}. Cultural differences significantly impact the style and content of communication in healthcare~\citep{kreuter-mcclure-2004-role,brooks-etal-2019-culturally} and is another area where much research is needed.
\citet{khoong-rodriguez-2022-research} further outline key domains for future research, including developing interactive tools for different types of communication; enhancing risk assessment, and assessing understanding and patient satisfaction on top of MT correctness.

\section{Conclusion}

Recontextualizing MT through Translation Studies and HCI highlights that truly supporting real-world needs demands understanding translation as a socio-technical process and designing user-centric tools. Each field offers important insights, and their synergy fuels new research.

Translation Studies provides theoretical and empirical frameworks for contextualizing assessments of translation quality, accounting for user diversity, and for framing translation as a process of situated decision-making that can inform our view of MT as it becomes part of increasingly diverse workflows.
HCI complements this by focusing on real-world user experience with translation technologies, emphasizing needs, interface design, feedback, and collaboration in multilingual interactions.
Both fields offer methods to evaluate stakeholder perceptions and behaviors, but mostly study the off-the-shelf MT and NLP tools which limits the space of interaction design. 
Conversely, MT/NLP offers a rich toolkit of generation, adaptation, and evaluation techniques, which are developed with less focus on user experience and context.

Interdisciplinary collaboration enables a shift towards genuinely human-centered systems, where users are active agents in a "machine in the loop" process.
This approach poses key technical challenges for MT: how to personalize translation outputs, how to support interaction and control, how to model trust and adaptation, how to balance generalization and responsiveness to context, and how to sustain human agency in language use.  Nevertheless, it promises greater real-world impact through more expansive conceptualizations of MT technology that support situated, embodied, and socially meaningful communication.

\section*{Limitations}

\paragraph{Scope} This survey is not exhaustive. While we aimed to highlight diverse perspectives, we cannot cover the breadth of the literature across Translation Studies, Human-Centered Interaction, Machine Translation and Natural Language Processing. To narrow down the scope, we facilitated discussions between experts in these disciplines to highlight connections and tensions across fields. We used the take-aways from these discussions to prioritize this survey. Furthermore, 
human-centered MT can draw upon insights and methodologies from many other disciplines, including linguistics and sociolinguistics, cognitive science and psychology, information science, communication studies, and education. 
\mc{expand a little on what each discipline offers, and add cites}

\paragraph{Multimodality} Most work surveyed here focused on text translation, but human-centered MT must incorporate multiple modalities, such as speech, images, and gestures, reflecting the way people communicate. In addition to speech translation \citep{akiba-etal-2004-overview,agarwal-etal-2023-findings} and its connection to simultaneous interpretation \citep{grissomii-etal-2014-don`t,wang-etal-2016-prototype}, prior work has considered the role of vision in translating image captions and video-guided translation \citep{specia-etal-2016-shared,sulubacak-etal-2020-multimodal}. Pre-trained language models that encompass speech \citep{radford-etal-2022-robust,ambilduke-etal-2025-tower} and vision \citep{radford-etal-2021-learningb,chen-etal-2024-scalinga} open new research directions. Further research with a human-centered perspective might include developing adaptive interfaces that detect errors \citep{han-etal-2024-speechqe}, seamlessly integrate multiple modalities and support repair \citep{sulubacak-etal-2020-multimodal}, thereby enabling more natural and effective human-computer interactions. However, a thorough treatment of multimodality in human-centered MT is beyond the scope of this paper.

\paragraph{Language Resource Disparities} Unequal coverage and quality of MT techniques across languages remains a fundamental limitation which must be taken into account to develop human-centered MT. Many methods discussed, particularly in Section~\ref{sec:hcmt}, are currently more feasible for high-resource languages. However, employing human-centered design methods and focusing on specific use cases can help develop strategies to mitigate disparities in translation quality across various languages, domains, and dialects \citep{santy-etal-2021-language}.

\section*{Acknowledgements}
This work was made possible by the NII Shonan Meeting on Human-Centered Machine Translation, organized by Marine Carpuat, Toru Ishida and Niloufar Salehi. We thank the National Institute of Informatics, Japan, for providing an excellent venue and support for productive discussions. We are also grateful to all participants for their contributions. We also thank Sharon O'Brien for earlier discussions.  

\bibliography{shonanreport,newrefs}

\end{document}